\documentclass[10pt,twocolumn,letterpaper]{article}

\usepackage{cvpr}
\usepackage{times}
\usepackage{epsfig}
\usepackage{graphicx}
\usepackage{amsmath}
\usepackage{amssymb}
\usepackage{subcaption}
\usepackage{booktabs}
\usepackage[pagebackref=true,breaklinks=true,letterpaper=true,colorlinks,bookmarks=false]{hyperref}

\cvprfinalcopy % *** Uncomment this line for the final submission

\def\cvprPaperID{1033} % *** Enter the CVPR Paper ID here

\ifcvprfinal\fi
\begin{document}

%%%%%%%%% TITLE
\title{ResFeats:  Residual Network Based Features for Image Classification}

\author{A. Mahmood$^{\star }$, M. Bennamoun, S. An\\
The University of Western Australia\\
{\tt\small $^{\star }$ ammar.mahmood@research.uwa.edu.au}
\and
F. Sohel\\
Murdoch University, Australia\\
}

\maketitle
%\thispagestyle{empty}

%%%%%%%%% ABSTRACT
\begin{abstract}
   Deep residual networks have recently emerged as the state-of-the-art architecture in image segmentation and object detection. In this paper, we propose new image features (called ResFeats) extracted from the last convolutional layer of deep residual networks pre-trained on ImageNet.  We propose to use ResFeats for diverse image classification tasks namely, object classification, scene classification and coral classification and show that ResFeats consistently perform better than their CNN counterparts on these classification tasks. Since the ResFeats are large feature vectors, we propose to use PCA for dimensionality reduction.  Experimental results are provided to show the effectiveness of ResFeats with state-of-the-art classification accuracies on Caltech-101, Caltech-256 and MLC datasets and a  significant performance improvement on MIT-67 dataset compared to the widely used CNN features. 
\end{abstract}

%%%%%%%%% BODY TEXT
\section{Introduction}

Deep convolutional neural networks (CNNs)  have shown outstanding results on  challenging image classification and detection datasets since the seminal work of \cite{krizhevsky2012imagenet}. Off-the-shelf image representations learned by these deep networks are powerful and generic. These generic features have been used to solve numerous visual recognition problems \cite{razavian2014cnn, donahue2014decaf}. Given the promising performance of these off-the-shelf CNN features, they have become the first choice for solving most computer vision problems \cite{azizpour2015generic}.

\begin{figure}[t]
\begin{center}
   \includegraphics[width=1\linewidth,height=1.8\linewidth, keepaspectratio]{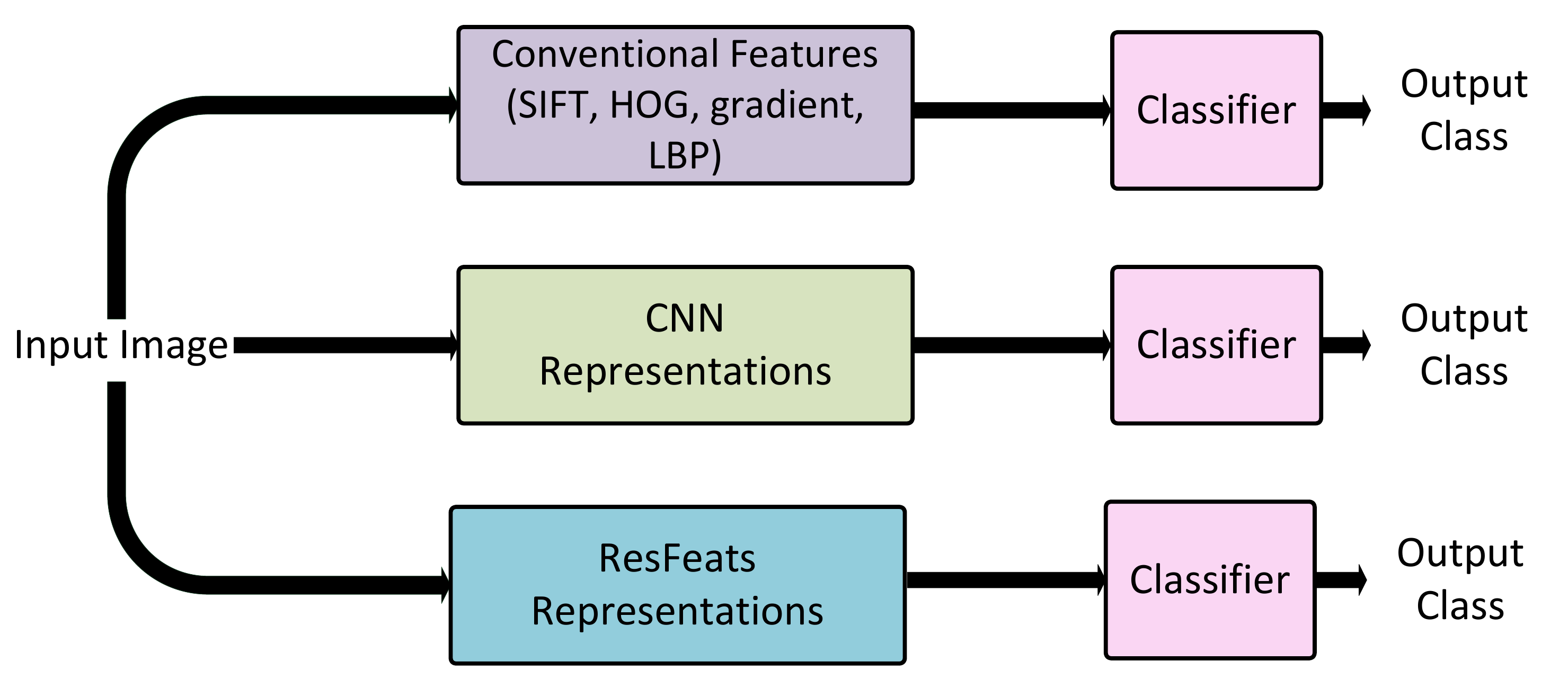}
\end{center}
   \caption{ Evolution of classification pipelines (the most recent an the bottom).  Off-the-shelf ResFeats have the potential to replace the previous classification pipelines and improve performance for image classification tasks.  }

\label{fig:feat}
\end{figure}

Training a deep network from scratch is not a feasible option when solving a classification problem with a small number of  labelled training examples. Recent evidence \cite{zeiler2014visualizing, he2014spatial, donahue2014decaf} suggests that off-the-shelf CNN features have outperformed previous handcrafted features for datasets with a limited amount of training data. These features are domain independent and can be transferred to any specific target task without compromising on performance \cite{azizpour2015generic}. Network width, depth and optimization parameters along with  the network layer from which these features are extracted play a key role in the effectiveness of transfer learning. This paper attempts to provide an answer to the following question: \textit{What are the criteria to select an initial deep network (pre-trained on ImageNet) to extract generic features in order to maximize performance and transferability across domains?} To answer this question, we hypothesise that a better optimized and a high performing  deep network on ImageNet should result in more powerful and generic image representations. One such network is the deep residual network (ResNet) presented in \cite{he2015deep}.
\begin{figure*}
\begin{center}
   \includegraphics[width=1\linewidth]{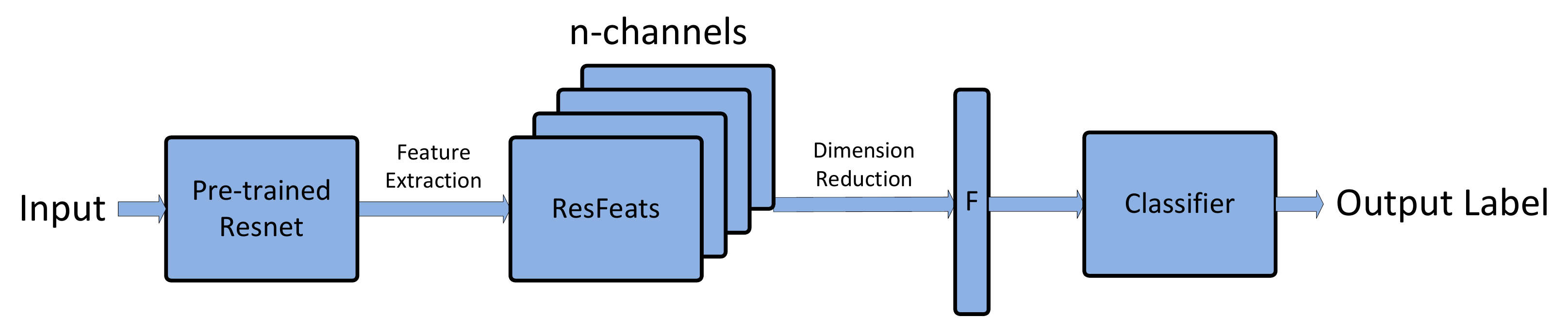}
\end{center}
   \caption{Block diagram of the proposed method. F is the final feature vector obtained after dimension reduction.}

\label{fig:block}
\end{figure*}

ResNets are easier to train as opposed to other CNN architectures \eg VGGnet \cite{simonyan2014very}. For example, a  152-layer ResNet which is 8 times deeper than VGGnet, is still less complex and trains faster.  Moreover, a 34-layer ResNet contains 3.6 billion multiply-add operations whereas a 19-layer VGGnet has 19.6 billion multiply-add operations (less than 20\%) \cite{he2015deep}. Very deep networks are known to cause overfitting and  saturation in accuracy. However, residual learning and the identity mappings (shortcut connections) \cite{he2016identity} in ResNets have been shown to  overcome these problems. This enables ResNets to achieve outstanding results in image detection, localization and segmentation tasks \cite{he2015deep}. In this paper, we explore the discrimination power of the image representations extracted from pre-trained ResNets. We name these off-the-shelf ResNet features as \textbf{ResFeats}.  Fig. \ref{fig:feat} depicts the evolution of traditional classification pipelines. 

The main contributions of this paper are listed below: 
\begin{itemize}
\item  We introduce ResFeats, which are image features extracted from pre-trained ResNets and test them on diverse image classification tasks including objects, scenes and corals. 
\item  We analyse the performance of  ResFeats extracted from the outputs of  different convolutional layers of ResNet-50 \cite{he2015deep} for image classification.  We also compare the performance of ResFeats extracted from ResNet-50 with those extracted from a deeper 152-layer ResNet.
\item We  propose a compact 2048-dimensional generic feature vector obtained after dimensionality reduction which is half of the size of the traditional CNN based feature vector (4096 dimensions). 
\item We show that ResFeats achieve a superior classification accuracy compared to off-the-shelf CNN features.  We also provide experimental evidence that our proposed method achieves state-of-the-art performance on three out of the four popular and challenging image classification datasets.
\end{itemize}

The rest of the paper is organized as follows: We briefly discuss the  related work in the next section. In Sec. 3.1, we introduce our proposed approach and explain the feature extraction from ResNets. In Sec. 3.2, we describe the dimensionality reduction and classification approaches. Sec. 4 reports the experimental results and Sec. 5 concludes the paper.

\section{Related Work}

 Recent success stories \cite{krizhevsky2012imagenet, simonyan2014very, donahue2014decaf, girshick2014rich} have established deep CNNs as the first choice to solve challenging computer vision tasks. However, training a network from scratch requires a large amount of training data, time and GPUs. Donahue \etal \cite{donahue2014decaf} and Zeiler and Fergus \cite{zeiler2014visualizing} provided  evidence that the generic image representations learned from pre-trained CNNs outperform previous state-of-the-art hand crafted features. However, they did not experiment on a large number of computer vision datasets. Razavian \etal \cite{razavian2014cnn} built on the concept of generic CNN  features and proved that off-the-shelf CNN features outperform existing methods. They experimented with more than 10 datasets for tasks such as image classification, object detection, fine grained recognition, attribute detection and visual instance retrieval.  OverFeat \cite{sermanet2013overfeat} was used as the source CNN in the work of \cite{razavian2014cnn}. 

Chatfield \etal \cite{chatfield2014return} evaluated the performance of CNN based methods for image classification and compared their methods with previous feature encoding methods. Their findings established that deeper CNN performed better than the shallower models of the same network trained on augmented data. VGGnet \cite{simonyan2014very} was used as the source CNN in their work. They improved the classification accuracies of popular datasets such as VOC, Caltech-101 and Caltech-256. He \etal \cite{he2014spatial} used spatial pyramid pooling of CNN features to further improve the classification accuracy on the Caltech datasets and reported state-of-the-art object classification results. 

Scene classification is quite different from object classification due to the presence of multiple objects in a single scene. These object instances can be of varying size and pose, and can be located at different locations in a number of possible layouts in the test image.   Consequently, the state-of-the-art performance on scene datasets such as MIT-67 (81\% in \cite{cimpoi2015deep}) is comparatively lower than the performance on object classification datasets (93.4\% for Caltech-101 in \cite{he2014spatial}).  Towards indoor scene classification, a bag of features approach was proposed to perform VLAD pooling \cite{jegou2010aggregating} of CNN features in \cite{gong2014multi}. Another example is ``spatial layout and scale invariant convolutional activations ($S^{2}ICA$)" introduced in \cite{7539697} to increase the robustness of CNN features.  Cimpoi \etal \cite{cimpoi2015deep} proposed Fisher Vector (FV) pooling of a deep CNN filter bank (FV-CNN) for texture and material classification. They achieved an accuracy of 81\% on MIT-67 dataset (an improvement of 10\% over previous state-of-the-art). 

Coral classification is a target task which is very different from the source dataset on which  deep networks are pre-trained (ImageNet in this case). Despite this dissimilarity, off-the-shelf CNN features have improved the results of existing methods of coral classification  \cite{mahmood2016coral, khan2015cost, mahmood2016oceans}, thereby demonstrating their strength for transfer learning.  The baseline performance on MLC dataset was first reported in \cite{beijbom2012automated}. In \cite{mahmood2016coral}, a hybrid (hand-crafted + CNN)  feature vector was proposed to improve the classification accuracy on this dataset. Khan \etal \cite{khan2015cost} used feature vectors extracted from VGGnet alongside cost-sensitive learning to address the class imbalance problem of MLC dataset. 
\begin{figure*}
\begin{center}
   \includegraphics[width=1\linewidth,height=1.5\linewidth, keepaspectratio]{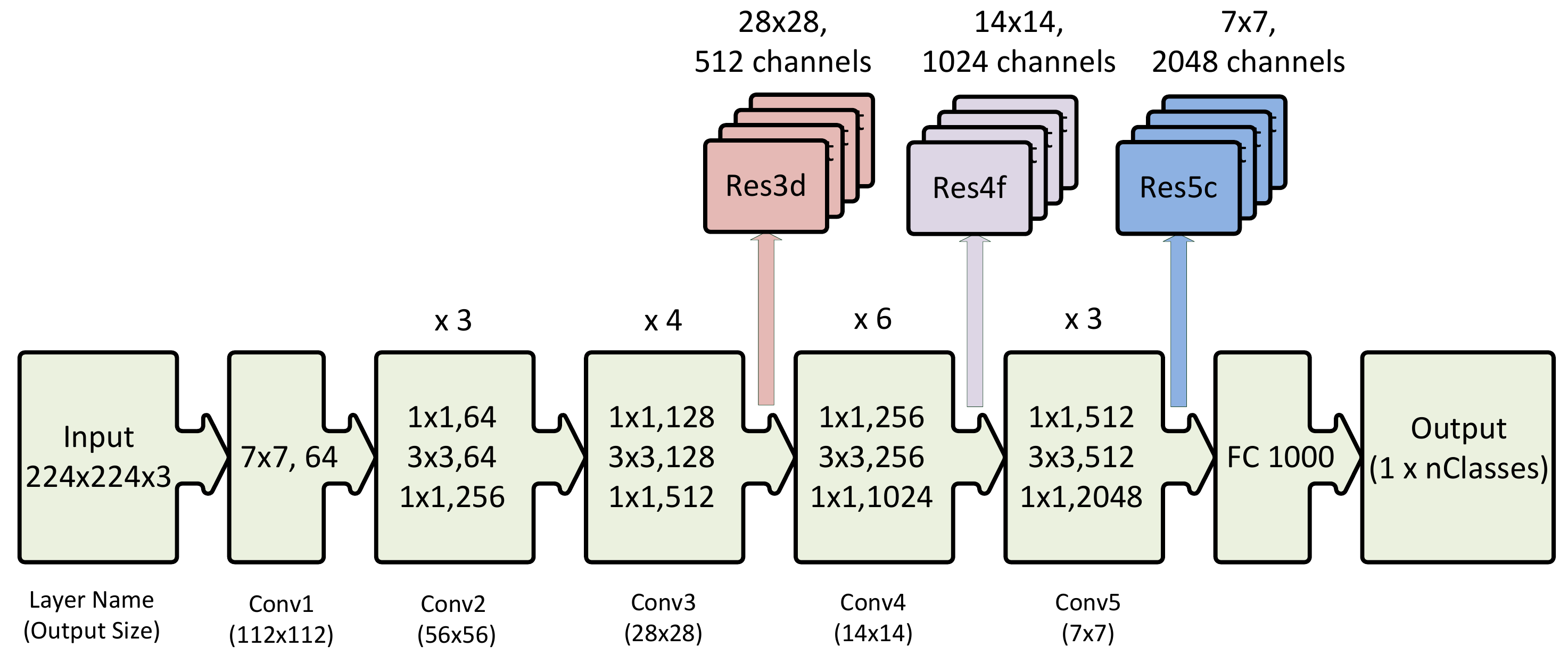}
\end{center}
   \caption{ResNet-50 architecture \cite{he2015deep} shown with the residual units, the size of the filters and the outputs of each convolutional layer. ResFeats extracted from the different layers of this network are also shown.}

\label{fig:res50}
\end{figure*}

\section{Proposed Method}
In the following subsections, we describe various steps that are involved in our proposed method with a block diagram in Fig. \ref{fig:block}.

\subsection{Deep Residual Networks}

Deep residual networks are made up of residual units. Each residual unit can be expressed as:

\begin{equation}
y_{i} = h(x_{i}) + \emph{F}(x_{i},w_{i})
\end{equation}

\begin{equation}
x_{i+1}= f(y_{i})
\end{equation}
where $\emph{F}$ is a residual function, $f$ is a ReLU function, $w_{i}$ is the weight matrix, and $x_{i}$ and  $y_{i}$ are the inputs and outputs of the $i$-th layer. The function $h$ is an identity mapping \cite{he2015deep} given by: 

\begin{equation}
h(x_{i})= x_{i}
\end{equation}

The residual function $\emph{F}$ is defined in \cite{he2016identity} as:

\begin{equation}
\emph{F}(x_{i},w_{i}) = w_{i} \cdot \sigma (B(w_{i}')\cdot \sigma(B(x_{i})))
\end{equation}
where $B(x_{i})$ is the batch normalization,``$\cdot$" denotes convolution and $\sigma(x)= max(x,0)$. The essential idea behind residual learning is the branching of the paths for gradient propagation. For CNNs, this idea was first introduced in the form of parallel paths in the inception models of \cite{szegedy2015going}. Residual networks share a few similarities with the highway networks \cite{srivastava2015highway} such as residual blocks and shortcut connections. However, the output of each path in the highway network is controlled by a gating function which is learned during the training phase.

The residual units in ResNets are not stacked together as is the case with convolutional layers in a conventional CNN.  Instead, shortcut connections are introduced from the input of each convolutional layer to its output. Using identity mappings as  shortcut connections decreases the  complexity of the residual networks resulting in deep networks that are faster to train. ResNets can be seen as an ensemble of many paths, instead of viewing it as a very deep architecture. However, all of these network paths in the ResNets are not of the same length. Only one path goes through all of the residual units. Moreover, all of these signal paths do not propagate the gradient which accounts for the faster optimization and training of ResNets. ResNets as deep as 1001-layers have been proposed to achieve superior performances on CIFAR datasets \cite{he2016identity}. However, in this paper we have only used ResNet-50 and ResNet-152 whose architectures are described in detail in \cite{he2015deep}.

\subsection{ResFeats}

This section introduces ResFeats and elaborates on the process to extract those features from deep residual networks. Generally, the image representations extracted from the deeper layers of a  CNN capture higher level features and increase the classification performance\cite{zeiler2014visualizing}. A typical residual unit in a ResNet consists of a block of three convolutional layers \cite{he2015deep}. ResFeats are the outputs of residual units unlike the conventional CNN features which usually are the activations of the fully connected layers \cite{razavian2014cnn}.  The activations of the fully connected layers capture the overall shape of the object contained in the region of interest. The local spatial information is lost when the outputs of the convolutional layer are max-pooled to obtain a 4096 dimensional vector for the activation of FC layer \cite{liu2015treasure}. However, the output vector of a convoltuional layer is rich in spatial information. 

ResFeats can be viewed as the output of a deep filter bank. This output is a vector of the form $ w \times h \times d$ where $w$ and $h$ is the width and height of the resulting feature vector and $d$ being the number of channels in the convolutional layer. Thus ResFeats can be considered as 2-D arrays of local features with $d$ dimensions.  The local spatial information of this feature vector will be lost when it is propagated to the fully connected layer. Therefore, we do not use the activations of the FC layer of ResNet as a feature vector.

Fig. \ref{fig:res50} shows the architecture of the ResNet-50 deep network which we have used for feature extraction. We initialize  the network with the weights pre-trained on ImageNet.  The learned weights of the deeper layers are usually  more class specific \eg the fully connected layer of ResNet-50 (since there is only one FC layer).  We were interested in the classification performance of the output vectors of the preceding convolutional layers. If used appropriately, the convolutional layers of a deep network form very powerful features. Therefore, we extracted the outputs of the last residual unit of the convolutional layers 3, 4 and 5 and used them as feature vectors. These feature vectors were denoted by Res3d, Res4f and Res5c respectively (the letters d, f and c correspond to 4 ,6 and 3 which is the number of the last residual blocks of each layer).  Features extracted from the 3rd layer have a lower dimension than the features extracted from the 5th layer. We expected an increase in the performance of ResFeats as we used deeper features. We also extracted these intermediate features from a deeper version of ResNet: ResNet-152 \cite{he2015deep}. ResNet-152 have shown a lower error on the ImageNet classification challenge than ResNet-50. Res5c features extracted from the 152-layer ResNet tend to perform better than their ResNet-50 counterparts. The classification results of these features are reported in Sec. 4.

\subsection{Dimensionality Reduction and Classification}

\begin{figure}[t]
\begin{center}
   \includegraphics[width=1\linewidth,height=1\linewidth, keepaspectratio]{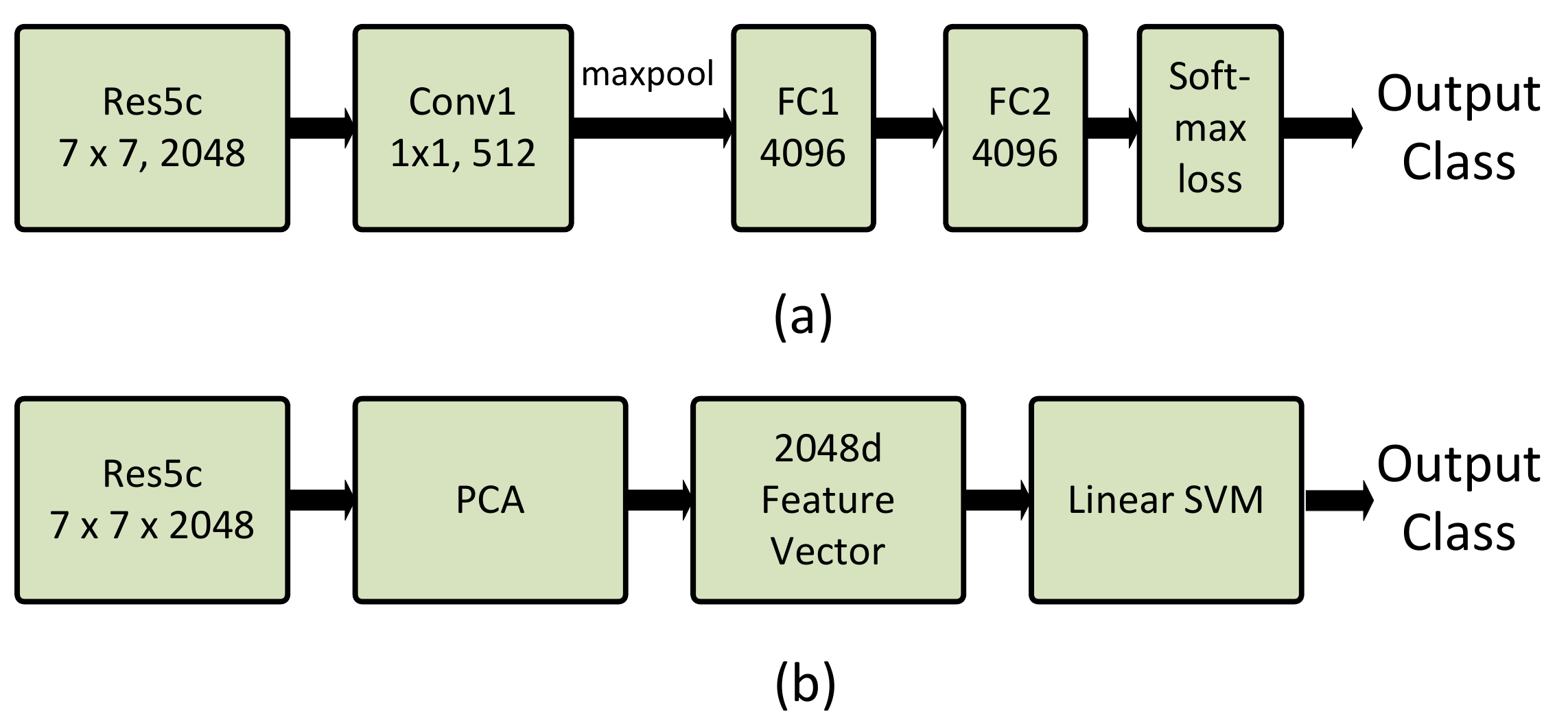}
\end{center}
   \caption{Dimension reduction and classification pipelines: (a) sCNN with two convolutional layers and two fully connected layers. (b) PCA-SVM. }

\label{fig:scnn}
\end{figure}

The outputs of the convolutional layers are much larger in size than the traditional 4096-dimensional CNN based features, for example, the Res5c feature vector is $7\times7\times2048$ in dimension (more than 100k elements). In order to reduce the computational costs associated with the manipulation of large feature vectors, we propose two methods for dimension reduction. The \textbf{first} method involves implementing a shallow CNN network with one convolutional layer, one max-pooling layer and two fully-connected (FC) layers. We will refer to this network as sCNN in the rest of the paper. The first convolutional layer consists of small filters (\ie $1\times1$)  along 512 channels. This layer reduces the dimension of Res5c to $7\times7\times512$ which is of the same size as the output of the last convolutional layer of VGGnet \cite{simonyan2014very}. The stride is set to 1 and the padding is set to zero for the convolutional layer.  This layer is then followed by a max-pooling layer, two FC layers and a soft-max layer for classification. The resulting shallow CNN is very similar to the FC portion of the VGGnet (configuration D \cite{simonyan2014very}). The resulting sCNN is initialized with random weights and is then trained for each dataset specifically. Fig. \ref{fig:scnn} (a) shows the architecture of sCNN along with the dimensions of the layers used for Res5c. 

In the \textbf{second} proposed method for dimension reduction, we use Principal Component Analysis (PCA) to reduce the Res5c feature vector to an $n$-dimensional vector. Here $n$ is the number of channels in the convolutional layer from which ResFeats are extracted. A validation set from each dataset is used to calculate the optimal $n$. The maximum validation accuracy is achieved when $n$ is set equal to the number of channels in the corresponding ResFeat. For example, Res5c ($7\times7\times2048$) is reduced to a 2048-dimensional vector by PCA. The resulting feature vectors are then classified using a linear support vector machine (SVM) classifier. We were motivated to use PCA-SVM classification pipeline due to its popularity to classify off-the-shelf CNN features \cite{azizpour2015generic, cimpoi2015deep, razavian2014cnn}. Fig. \ref{fig:scnn} (b) shows the pipeline for PCA-SVM module for Res5c.  A comparison of the performance of these two methods is also given in Sec. 4. Our results show that the dimensionality of the ResFeats can be reduced significantly without having a considerable performance drop. 

\section{Experiments and Results}

\subsection{Datasets}

\textbf{ Object Classification: Caltech-101} \cite{fei2006one} contains 9,144 images, divided into 102 categories. The number of images for each category varies between 31 and 800 images. In our experiments, we used 30 images from each class for training and the remaining images were used for testing. Caltech-101 is a very popular dataset for object classification.

\textbf{ Object Classification: Caltech-256} \cite{griffin2007caltech} contains 30,607 images, divided into 257 classes (256 objects +1 background). Each category has at least 80 images. This dataset is less popular but more challenging compared to Caltech-101. In our experiments, following \cite{zeiler2014visualizing}, we used 30 and 60 images from each class for training and the rest of the images were used for testing. 

\textbf{Scene Classification: MIT-67} \cite{quattoni2009recognizing} is a very challenging and popular dataset for indoor scene classification. It consists of 15,620 images belonging to 67 classes. The number of images varies between 101 and 738 per class. We followed the  standard protocol \cite{quattoni2009recognizing} which uses a subset of 6700 images (100 per class) for training and testing. There are 80 images from each class in the training set. The remaining 20 images per class are set for testing.  We also tested on the augmented version of this dataset by adding cropped and rotated samples. We refer it to as 'MIT-67aug' in our results. 

\textbf{ Coral Classification: Moorea Labelled Corals (MLC)} \cite{beijbom2012automated} contains 2055 images collected over three years: 2008, 2009 and 2010. It  contains random point annotation \textit{(x, y, label)} for the nine most abundant labels, four non coral and five coral classes. We have used 87,428 images from the year 2008 for training and the remaining 43,832 images from the same year for testing. This  is a challenging dataset since  each class exhibits a large variability in shape, color and scale. 

\subsection{Experimental Settings}

We use two deep ResNets to learn our proposed  image representations. The network architecture of the first ResNet is shown in Fig. \ref{fig:res50}. The detailed achitecture of the much deeper ResNet152 is similar to ResNet-50 and is illustrated in detail in \cite{he2015deep}. We use the pre-trained models of these two networks which are publicly available. We implemented our proposed method and sCNN classifier network in MatConvNet\cite{vedaldi15matconvnet}. LibSVM \cite{CC01a} was used for training the support vector machines used for classification. $n$-fold cross validation was used to find the best parameters for SVM with $n=4$. Note that the PCA-SVM was only tested for the highest performing ResFeats \ie , ResFeats-152.

The classification accuracies reported in Sec. 4.3 and 4.4 were achieved by using the sCNN for dimensionality reduction and classification. A performance  comparison between sCNN and PCA-SVM module is given in Sec. 4.5 for ResFeats extracted from ResNet-152.

\subsection{Performance Analysis:  ResFeats }

In Table \ref{table:res50}, we present the classification accuracies of ResFeats extracted from the output of the 3rd, 4th and 5th convolutional layers on our test datasets. ResFeats from the 5th convolutional layer (Res5c) outperform others for all datasets except the MLC. The difference in the classification accuracy of the ResFeats extracted from different layers tends to follow a pattern that can be associated with the number of classes in the dataset. When the number of classes increases, the difference in the accuracies of Res5c, Res4f and Res3d also increases. For Caltech-256 (257 classes), the difference in the accuracy of Res5c and Res3d ranges between 30-35\%.  This difference is negligible for MLC dataset which only has nine classes. We conclude that high level features (\ie Res5c) show the best performance on all datasets except MLC. The same pattern was observed for the corresponding features extracted from ResNet-152. 

\begin{table}
\begin{center}
\scalebox{0.9}{
\begin{tabular}{ @{} ccccc @{}  }
\toprule
Dataset & Classes& Res5c & Res4f & Res3d \\
\midrule
Caltech 101 (30) &102 & \textbf{91.8} & 89.4 &	77.2 \\
Caltech 256 (30)  & 	257 &\textbf{75.4} &	45.2 &46.0 \\
Caltech 256 (60)  &257 & \textbf{79.3} & 	53.4 & 44.1\\
MIT-67	&67&\textbf{71.1	}& 69.0 & 51.4\\
MLC 	&9&  76.8  & 	\textbf{78.8} & 77\\
\bottomrule
\end{tabular} }
\end{center}
\caption{Performance comparison of ResFeats extracted from different convolutional layers of ResNet-50. The number in the parenthesis denotes the number of samples per class that is used for training.} \label{table:res50}
\end{table}

\subsection{Performance Analysis: CNN features vs ResFeats}
Table \ref{table:cnn} compares the performance of ResFeats with their CNN counterparts for a given dataset. The overall classification accuracy is used to evaluate the performance. To keep the comparison fair, standard train-test splits are used for all datasets. For a fair comparison of classification performance, we only consider the methods which have used CNN features without any post-processing. We compare the CNN features with ResFeats extracted from a 50-layer ResNet and a deeper 152-layer ResNet. ResFeats-50 consistently outperform the CNN features by a margin of at least 4\%. Table \ref{fig:scnn} also shows that ResFeats-152 further improves the classification accuracy by 1-2\%.   We conclude that ResFeats perform significantly better than the corresponding CNN based features. Moreover, ResFeats extracted from a deeper ResNet perform better than the ones extracted from shallower ResNets.

\begin{table}
\begin{center}
\scalebox{0.9}{
\begin{tabular}{ @{} cccc @{}  }
\toprule
Dataset & CNN Features & ResFeats-50 & ResFeats-152 \\

\midrule
Caltech 101 (30) & 86.5 \cite{zeiler2014visualizing} & 91.8 &	\textbf{92.6} \\
Caltech 256 (30)  & 	70.6 \cite{zeiler2014visualizing}  &	75.4 & \textbf{78.0} \\
Caltech 256 (60)  & 74.2 \cite{zeiler2014visualizing}  & 	79.3 & \textbf{81.9}\\
MIT-67	& 58.4 \cite{razavian2014cnn}	& 71.1 & \textbf{73.0}\\
MIT-67aug	& 69.0 \cite{razavian2014cnn}		& 73.0 & \textbf{74.0}\\
MLC 	& 72.9 \cite{khan2015cost}  & 	78.8 &	\textbf{80.0}\\
\bottomrule
\end{tabular} }
\end{center}
\caption{Performance comparison of the baseline CNN features with the baseline ResFeats without any additional post-prcessing of feature vectors. The number in the parenthesis denotes the number of samples per class  that is used for training.} \label{table:cnn}
\end{table}

\subsection{Image Classification Results}
The experiments above compare our ResNet based feature representation with off-the-shelf CNN features. In this section, we  compare the performance of ResFeats with other state-of-the-art methods for each dataset. 

\textbf{Caltech-101:} We randomly select 30 images per class for training and compare our results with the other existing methods in Table \ref{table:cal}. ResFeats with a PCA-SVM classifier beats the current state-of-the-art (He \etal \cite{he2014spatial}) by 1.3\%. It is worth mentioning here that the authors in \cite{he2014spatial} used the spatial pyramid pooling layer in their network to achieve a 93.4\% accuracy. We, however, have achieved state-of-the-art accuracy without adding any post-processing modules to ResFeats. This demonstrates the superior classification power of ResFeats. 

\begin{table}
\begin{center}

\begin{tabular}{ @{} cc @{}  }
\toprule
Method & Cal-101 (30)  \\

\midrule
Bo \etal	\cite{bo2013multipath} & 81.4  \\
Zeiler \& Fergus \cite{zeiler2014visualizing} & 	86.5  \\
Chatfield \etal \cite{chatfield2014return}	& 88.4\\
He \etal \cite{he2014spatial}	& 93.4 	\\
\midrule
ResFeats-50 + sCNN	& 91.8	\\
ResFeats-152 + sCNN	& 92.6 	\\
ResFeats-152 + PCA-SVM	& \textbf{94.7} \\
\bottomrule
\end{tabular} 
\end{center}
\caption{Performance evaluation on Caltech-101 dataset. The number in the parenthesis denotes the number of samples per class  that is used for training. } \label{table:cal}
\end{table}

\textbf{Caltech-256:} We randomly select 30 and 60 images per class for training and report the classification accuracies in  Table \ref{table:cal256}. Our method (both classification modules) outperforms the current state-of-the-art in both experiments. Table. \ref{table:cal256} reports an absolute gain of 8.9\% and 4.5\% on previous state-of-the-art methods on Caltech-256 datasets with 30 and 60 training samples per class respectively.

\textbf{MIT-67:}  We report our results on the standard split (80 train, 20 test)  on MIT-67 and the augmented version (MIT67-aug) of this dataset in Table \ref{table:mit}. We use 16 augmentations of each image: five crops, two rotations and mirrored images of these. The data augmentation used in our experiments is consistent with the one used in \cite{razavian2014cnn}. Table \ref{table:mit} shows that ResFeats perform better than all the  previous methods except \cite{cimpoi2015deep} for the non-augmented dataset. The best performing method on MIT-67, Cimpoi \etal used deep filter banks that are extracted from VGGnet at multiple scales followed by a Fisher Vector (FV) encoding to achieve state-of-the-art performance on MIT-67. However, it is important to note that applying FV encoding to ResFeats is computationally expensive because of the large size of ResFeats (Res5c has more than 100k elements). Also, this method extracted features from the last layer convolution layer of VGGnet by using multiple sizes of each training image. In contrast, we only use a fixed size ($224\times224$) to extract ResFeats.  For MIT67aug, our method beats the previous best performance by a margin of 8.1\%. 

\textbf{MLC:}  We use the same experimental protocol for MLC dataset as given in \cite{beijbom2012automated}. Table \ref{table:mlc} shows the classification accuracies for MLC dataset achieved by previous methods. Our proposed method achieves an accuracy gain of 6.8\% over the baseline performance of \cite{beijbom2012automated}. Off-the-shelf ResFeats outperform the cost-sensitive CNN of \cite{khan2015cost} and multi-scale hybrid feature (CNN + hand-crafted feature) approach of \cite{mahmood2016coral}.

\begin{table}
\begin{center}

\begin{tabular}{ @{} ccc @{}  }
\toprule
Method &  Cal-256 (30) & Cal-256 (60) \\

\midrule
Sohn \etal \cite{sohn2011efficient} & 42.1 & 47.9 \\
Bo \etal	\cite{bo2013multipath}  &48.0 &	55.2 \\
Zeiler \& Fergus \cite{zeiler2014visualizing}  &	70.6 &	74.2 \\
Chatfield \etal \cite{chatfield2014return}	&  	-- & 77.6\\
\midrule
ResFeats-50 + sCNN	 	& 75.4	 & 79.3\\
ResFeats-152 + sCNN	 	& 78.0	 & 81.9\\
ResFeats-152 + PCA-SVM	 & 	\textbf{79.5} &	\textbf{82.1}\\
\bottomrule
\end{tabular} 
\end{center}
\caption{Performance evaluation on Caltech-256 dataset. The number in the parenthesis denotes the number of samples per class  that is used for training. } \label{table:cal256}
\end{table}

\begin{table}
\begin{center}
\begin{tabular}{ @{} ccc @{}  }
\toprule
Method & MIT-67 & MIT-67 aug \\

\midrule
Razavian \etal	\cite{razavian2014cnn} & 58.4 &69.0  \\
Gong \etal \cite{gong2014multi} & 	68.9 &	--  \\
Khan \etal \cite{khan2015cost}	& 70.9 & 	-- \\
Zhou \etal \cite{zhou2014learning}	& 70.8 	& --	 \\
Azizpour \etal \cite{azizpour2015generic}	& 71.3 	& --	 \\
Liu \etal \cite{liu2015treasure} & 71.5 & --\\
Hayat \etal \cite{7539697} & 74.4 & --\\
Cimpoi \etal \cite{cimpoi2015deep}	& \textbf{81.0} 	& --	 \\
\midrule
ResFeats-50 + sCNN classifier	 	& 71.1 & 	73.0\\
ResFeats-152 + sCNN classifier	 	& 73.7 & 	74.9\\
 ResFeats-152 + PCA-SVM	 	& 75.6 & 	\textbf{77.1} \\
\bottomrule
\end{tabular}
\end{center}
\caption{Performance evaluation on MIT-67 dataset. } \label{table:mit}
\end{table}

\begin{table}
\begin{center}
\begin{tabular}{ @{} ccc @{}  }
\toprule
Method & MLC  \\

\midrule
Beijbom \etal \cite{beijbom2012automated}	& 74.0 		 \\
Khan \etal \cite{khan2015cost}	& 75.2 \\
Mahmood \etal \cite{mahmood2016coral}	& 77.9	 \\
\midrule
ResFeats-50+ sCNN classifier	 	& 78.8 \\
ResFeats-152 + sCNN classifier 	& 80.0  \\
ResFeats-152 + PCA-SVM	& \textbf{80.8}  \\
\bottomrule
\end{tabular}
\end{center}
\caption{Performance evaluation on MLC dataset. } \label{table:mlc}
\end{table}

\begin{figure*}
\begin{center}
   \includegraphics[width=1\linewidth,height=1\linewidth, keepaspectratio]{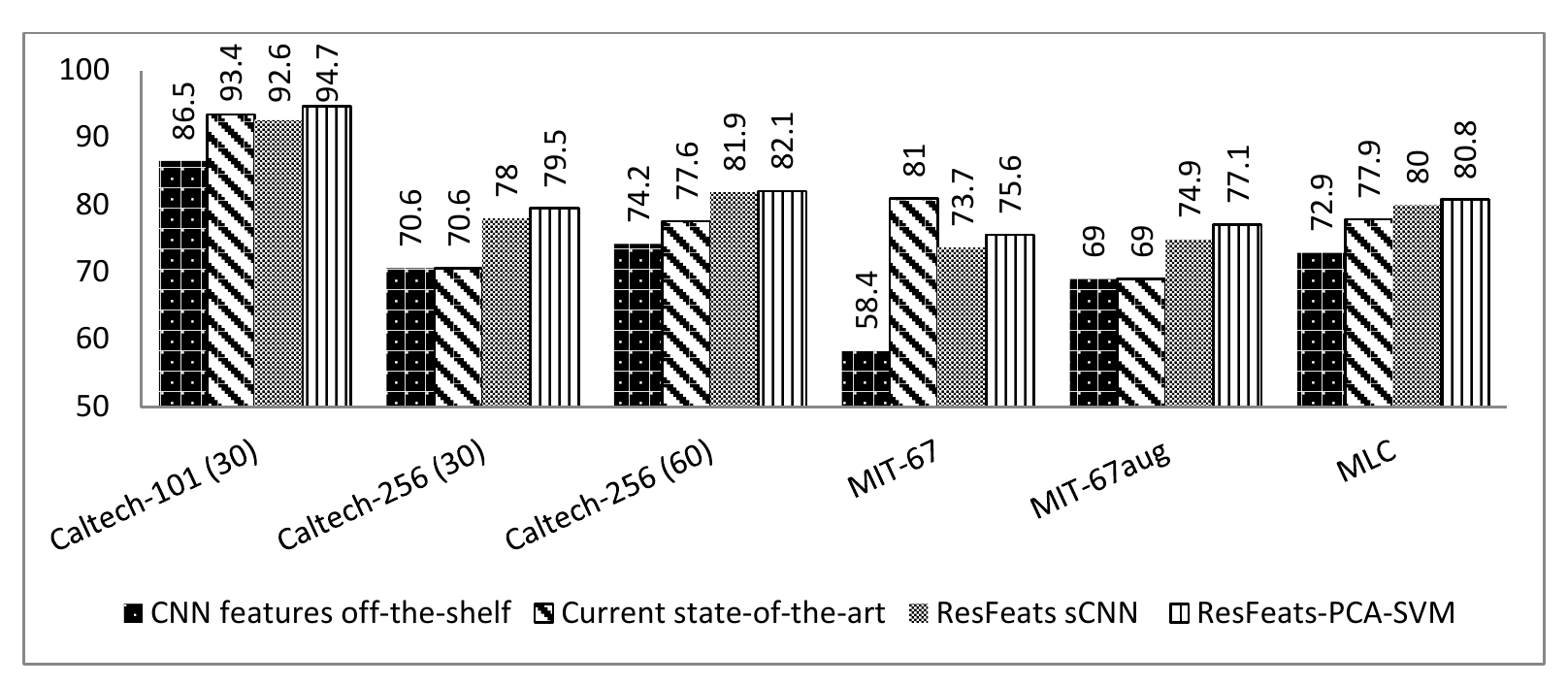}
\end{center}
   \caption{The improvement achieved by replacing CNN off-the-shelf features with ResFeats for the datasets we used in our experiments. Current state-of-the-art performances are also given for each dataset.}
\label{fig:exc}
\end{figure*}

Fig. \ref{fig:exc} shows a comparison of the classification accuracy of off-the-shelf CNN representations, ResFeats and current state-of-the-art methods. The results are reported for all the datasets that were used in our experiments. The ResFeats consistently outperformed the CNN features by a large margin. It must be noted in Fig. \ref{fig:exc} that for CNN features, only those results are reported which do not use any additional post-processing module.  ResFeats with PCA-SVM achieved state-of-the-art classification performances for all the datasets except MIT-67.

\section{Conclusion}

In this paper, we used features extracted from deep ResNets off-the-shelf to address three image classification tasks: object, scene and coral classification. We investigated the effectiveness of  transfer learning of  the ResFeats. We showed that the ResFeats extracted from the deeper layers of a ResNet perform better than the shallower ResFeats.  We experimentally confirm that our proposed features are powerful and have a classification accuracy that is higher than the CNN off-the-shelf features. Finally, we improve the state-of-the-art accuracy on Caltech-101, Caltech-256 and MLC datasets. It is worth to further investigate  the prospective applications of ResFeats for computer vision tasks such as object localization, image segmentation, instance retrieval and attribute detection. 

\section{Acknowledgements}

We thankfully acknowledge Nvidia for providing us a Titan-X GPU for the experiments involved in this research. 
%-------------------------------------------------------------------------

{\small
\bibliographystyle{ieee}
\bibliography{egbib}
}

\end{document}